\documentclass[12pt]{article} 
\usepackage{amsmath,amsthm,amssymb,bm,mathrsfs}
\usepackage{graphicx,subfigure}
\usepackage{caption,enumerate,listings,setspace}
\usepackage[colorlinks,linkcolor=red,anchorcolor=blue,citecolor=blue]{hyperref}
\usepackage[margin=1in]{geometry}
\usepackage[title]{appendix}
\usepackage{enumitem}
\theoremstyle{plain}
\usepackage{booktabs}
\usepackage[numbers]{natbib}
\usepackage{array}
\usepackage{algorithm}
\usepackage{algpseudocode}
\usepackage{makecell}

\let\tilde\widetilde
\numberwithin{example}{section}

\usepackage[utf8x]{inputenc}
\usepackage{mathrsfs}
\usepackage{amssymb}
\usepackage{amsfonts}
\usepackage{amsmath}
\usepackage{amsthm,verbatim}
\usepackage{wrapfig}
\usepackage{multirow}
\usepackage{longtable}
\usepackage{enumerate}
\usepackage{color, graphicx}
\usepackage{tikz}

\def\boxit#1{\vbox{\hrule\hbox{\vrule\kern6pt
          \vbox{\kern6pt#1\kern6pt}\kern6pt\vrule}\hrule}}
\title{\textbf{Data Plagiarism Index: Characterizing the Privacy Risk of Data-Copying in \\ Tabular Generative Models}} 

\author
{ Joshua Ward \thanks{Ph.D. Candidate, Department of Statistics and Data Science,  UCLA, CA, 90095. Email: joshuaward@ucla.edu}, 
Chi-Hua Wang \thanks{Postdoctoral Scholar, Department of Statistics and Data Science,  UCLA, CA, 90095. Email: chihuawang@ucla.edu},
Guang Cheng\thanks{Professor, Department of Statistics and Data Science, UCLA, CA, 90095. Email: guangcheng@ucla.edu}
}

\begin{document} 

\maketitle

\begin{abstract}
The promise of tabular generative models is to produce realistic synthetic data that can be shared and safely used without dangerous leakage of information from the training set. In evaluating these models, a variety of methods have been proposed to measure the tendency to copy data from the training dataset when generating a sample. However, these methods suffer from either not considering data-copying from a privacy threat perspective, not being motivated by recent results in the data-copying literature or being difficult to make compatible with the high dimensional, mixed type nature of tabular data. This paper proposes a new similarity metric and Membership Inference Attack called Data Plagiarism Index (DPI) for tabular data. We show that DPI evaluates a new intuitive definition of data-copying and characterizes the corresponding privacy risk. We show that the data-copying identified by DPI poses both privacy and fairness threats to common, high performing architectures; underscoring the necessity for more sophisticated generative modeling techniques to mitigate this issue. 
\end{abstract}


\clearpage
\section{Introduction}
\label{sec:intro}

Data-copying is a particularly concerning manifestation of generative models overfitting \cite{meehan2020three, bhattacharjee2023data}. Historically, researchers have observed that deep learning models exhibit overfitting behaviors, sometimes generating unrealistic instances and at other times creating instances overly similar to samples in the training dataset. While the former can be lauded as "creative and imaginative," the latter poses a significant risk, threatening the confidentiality of training data. Indeed, previous studies \cite{alaa2022faithful} reveal that many prominent generative models attain high scores in terms of fidelity and diversity by memorizing or copying real samples, which compromises their effectiveness for privacy-sensitive applications.
The issue of data-copying is paramount in the field of tabular generative models, especially as these models are often used in scenarios involving sensitive data and strict privacy protocols \cite{yoon2020anonymization}. This paper focuses on the challenge of detecting and evaluating data-copying in tabular data generation, highlighting its critical role in enhancing the trust and accountability of these methods.

\begin{figure}[H]
    \centering
    \includegraphics[width=\linewidth]{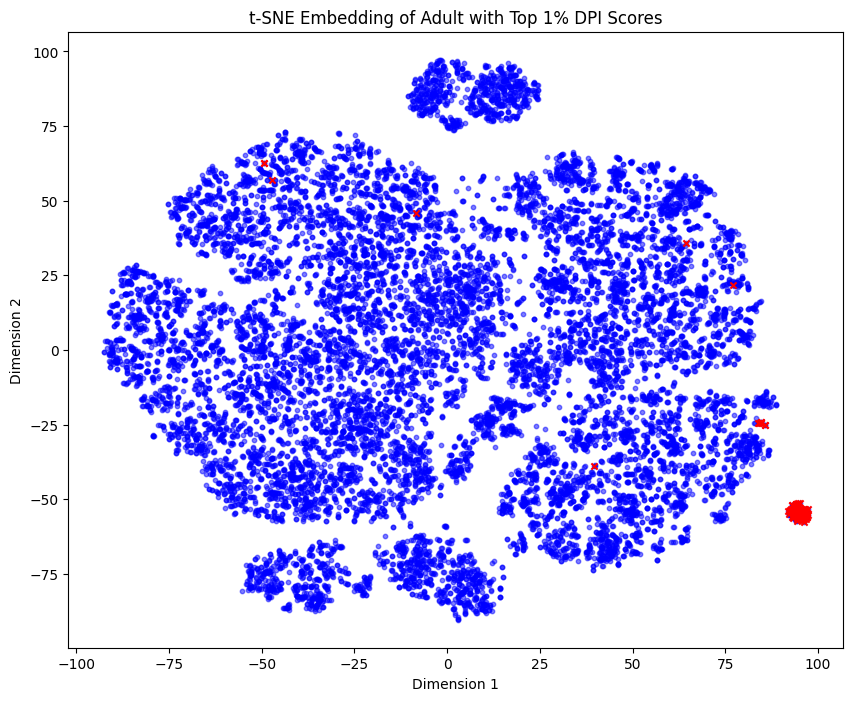}
    \caption{A t-SNE plot of Tab-DDPM's training data with corresponding top 1\% DPI Scores in red on the Adult dataset. DPI identifies an outlier region in the bottom right corresponding to an extreme privileged class (married, white, middle aged, high capital gains, private industry, respondents making $>$50k in income). This provides evidence that Tab-DDPM copies the training data of outlier and privileged classes, creating serious fairness concerns for practitioners who use synthetic data in their downstream machine learning tasks. 
    See Sec. \ref{subsec:Train_Data_High_DPI} for detailed discussions.}
    \label{fig:tsne}
\end{figure}

Various paradigms have been proposed to study data-copying in generative models, including hypothesis testing  \cite{meehan2020three}, non-parametric statistics \cite{bhattacharjee2023data}, Membership Inference Attacks \cite{vanbreugel2023membership}, and ad-hoc similarity metrics \cite{platzer2021holdout, alaa2022faithful, solatorio2023realtabformer}. Each of these approaches contributes unique insights into the phenomenon of data-copying in tabular generative models but have some sort of drawback. For instance, while similarity metrics used in tabular synthetic data literature offer an understanding of the geometric relationship between training and generated data, they lack a threat model to assess the \textit{privacy risks} associated with these geometries (see \cite{ganev2023inadequacy}). Conversely, Membership Inference Attacks provide valuable tools for understanding privacy risks but are often disconnected from the model overfitting literature and are challenging to apply to the complex, high-dimensional and mixed-type distributions that are common in tabular data applications. The disjoint nature of the data-copying literature fails to address common practitioner questions, such as: 
\begin{center}
\textit{'To what extent does a model copy training data, and how practically significant is this problem?'}
\end{center}
Unfortunately, we show that the problem is substantial, as evidenced by Figure \ref{fig:tsne}. 

In this paper, we study data-copying in tabular generative model from the perspective of all three of these disconnected areas (Membership Inference Attacks, ad-hoc similarity metrics and Data-Copying measures). The goal is to craft a principled, interpretable, and privacy orientated metric that can be applied to tabular generative modelling. Here, we propose Data Plagiarism Index (DPI); a theory-motivated measure of local data-copying (Section \ref{subsec:Data_Plag_Index}). We argue that DPI differs from other competing metrics and show how it provides a novel geometric perspective on the data-copying behavior of generative models. With the proposed privacy metric DPI, we further develop a new type of Membership Inference Attack, named DPI MIA (Section \ref{subsec:DPIMIA}), that bridges this data-copying measure with testable privacy risk. Our experiment (Section \ref{sec:results}) find that DPI MIA identifies a tendency among high-fidelity tabular data generators to perform risky data-copying, raising privacy concerns.

Furthermore, our empirical evidence (Figure \ref{fig:tsne}) reveals that tabular generative models \textbf{disproportionately favor certain privileged sub-populations} within the synthetic sample of the classic Adult dataset, raising serious fairness concerns when practitioners use synthetic data in downstream machine learning tasks. This troubling finding underscores that the proposed Data Plagiarism Index (DPI) is not merely a technical curiosity but a critical issue that can compromise both the privacy and fairness of tabular generative models.

Overall, we summarize our contribution to the Trustworthy Generative Modeling literature as follows:
\begin{enumerate}[leftmargin=*]
    \item We propose a novel privacy measure called \textbf{Data Plagiarism Index (DPI)} and a corresponding Membership Inference Attack called DPI MIA that measures the privacy risk of synthetic tabular data from data-copying.
    \item We provide empirical evidence (Figure \ref{fig:model_perf_v_mia}) that DPI identifies a positive correlation between tabular generative models' utility and privacy risks. 
    \item We provide empirical evidence (Figure \ref{fig:tsne}) that reveals that Tab-DDPM \cite{tabddpm} drastically copies the source training data of privileged sub-classes, creating a source of structural unfairness in the synthetic data. 
    \item We show (Figure \ref{fig:model_perf_v_mia}, Table \ref{tab:pcor}) that DPI MIA can identify a different kind of data-copying undetectable by existing Membership Inference Attacks with comparable threat performance, providing a new way to deploy privacy attacks to audit synthetic data privacy. 
\end{enumerate}

% %\clearpage
\section{Related Work}
\label{sec:related_works}

In this section, we review three major literature to measure the overfitting and data-copying phenomenon of tabular generative models.

\subsection{Measure Data-Copying in Generative Models}
Generative models' data-copying in the literature is identified when synthetic data is excessively similar to training samples \cite{meehan2020three}. This is typically evaluated using a reference dataset, sampled from the same distribution as the training dataset. The common method involves an "appropriate distance function" to check whether synthetic data is closer to the training data or the reference data \cite{platzer2021holdout}. \cite{meehan2020three} for example focuses on identifying a suitable distance function to determine whether a synthetic dataset or a reference dataset is closer to the training dataset, thus testing if generative models exhibit data-copying. They devised a non-parametric method that divides the instance space into cells, tests each cell individually, and combines the results to understand the overall degree of data-copying. This concept inspired our more sophisticated data-copying measure (See Section \ref{subsec:data_copy_measure}). \cite{bhattacharjee2023data} introduces a more advanced data-copying test, capable of detecting data-copying behaviors not identified by \cite{meehan2020three}. Both of these studies however have limitations in high-dimensional complex distributions and also do not evaluate data-copying from a privacy perspective. DPI is more practical and accurately captures the degree of local data-copying, aiding in the assessment of synthetic data privacy risks (See Fig \ref{fig:tsne}).

\subsection{Similarity Metrics between Real and Synthetic Data}
\label{subsec:ref_similarity_metrics}

A variety of ad-hoc metrics have been proposed to evaluate the privacy of tabular synthetic data from a model overfitting perspective. Broadly speaking, these metrics focus on determining the level of similarity between the training and synthetic datasets ideally hoping to find that the generated data is in a 'Goldilocks' zone: not too similar to the training data, but also not too dissimilar. These metrics will often be posed as a comparison between the training and a reference set and training and synthetic sets creating a sort of 'Null Distribution' in which to test privacy.  

A common example of an often used similarity metric is Distance to Closest Record (DCR) \cite{park2018data, lu2019empirical, yale2019assessing, zhao2021ctab,guillaudeux2023patient, liu2023tabular} where for each training point the distance to its closest neighbor in the synthetic dataset is compared with the distance of the closest neighbor in the reference dataset. Another is Identical Matching Score \cite{ims1, ims2, lu2019empirical} which compares the proportion of identical records in the training and synthetic observations.

There are a variety of problems with this style of privacy evaluation. The first is that these similarity metrics do not actually guarantee nor are they evaluated by any idea of privacy protections against some sort of attack. For example, passing or failing a DCR or IMS test does not provide any information for how well a privacy attack may or may not do. Secondly, posing privacy as a hypothesis testing problem is a cardinal Statistics sin in that if the null hypothesis is: "the synthetic data are private" and the alternative is: "the synthetic data are not private", passing the test does not imply confirmation of the null, only that there is a failure to find sufficient evidence in which to reject the null \cite{CaseBerg:01}. However, while not being theoretically motivated, these metrics do provide an intuitive tool to study the geometric relationships between the training and synthetic datasets. Technically, DPI can be classified as a similarity metric and used in this way but these problems motivate DPI to adopt a Membership Inference Attack paradigm to overcome these dubious privacy interpretations.

\subsection{Membership Inference Attacks for Generative Models}
\label{subsec:ref_MIA}

Membership Inference Attacks have traditionally been studied in a supervised learning context and only relatively recently have become applied to generative models. Here, the goal of the attack is to determine whether some sample of test data $x^* \in X^*$ was used in the training of the model based on some information about that model \cite{shroki}. The scenario of what information is available for an attacker is called the threat model to which there are a variety of identified contexts for generative model MIAs. These include black box attacks \cite{Hayes2017LOGANMI, Hilprecht2019MonteCA, ganleaks} in which only generated synthetic data is available, white box attacks in which both synthetic data and the internals of a model are known, and calibrated (also called shadow) attacks in which both a synthetic dataset and then a reference dataset from the same or similar distribution as the training set are given \cite{ganleaks, Hayes2017LOGANMI, vanbreugel2023membership}. Most attacks generally fall into the black box and calibrated paradigms as white box attacks are usually specific to the model architecture in question \cite{sablayrolles2019white}. The value of MIAs is that they provide a tangible, practical scenario in which to study privacy risks. We will later frame DPI as an MIA in order to study how identifying local data-copying corresponds to privacy leakage.

\section{Preliminaries}

\subsection{Formal definition of Data-Copying}
\label{subsec:data_copy_measure}

We first introduce a formal definition of data-copying motivated by \cite{meehan2020three} to provide theoretical context for Data Plagiarism Index. 
Given a data distribution $\mathbb{P}$ and a region $C$, consider the following probability measure: $\mathbb{P}|_{C}(A) \equiv \mathbb{P}(A \cap C)/\mathbb{P}(C)$ for all measurable set $A$. 
Let $R$ denote the distribution of reference data points and $S$ denote the distribution of synthetic data points. 

Given a neighborhood $D(x)$ of target data point $x$, define a one-dimensional distribution by $L(R) \equiv I(R \in D(x))$  and $L(S)\equiv I(S \in D(x))$ to denote separately the event of the reference data distribution and synthetic data distribution belonging to the neighborhood $D(x)$. By definition, $L(R) \sim R|_{D(x)}$ and $L(S) \sim S|_{D(x)}$

Define $\Delta(R, S)=P(B>A| B\sim L(S), A\sim L(P)$ to be the event that the synthetic data points have a greater probability in belonging to neighborhood $D(x)$ than the reference data points. 

In the spirit of Definition 2.1 of \cite{meehan2020three}, we define a generative model as \textbf{data-copying} a training data point $x$, if there exists a neighborhood $D(x)$ of $x$ such that the synthetic data distribution $S$ is systematically closer to the training data point $x$ than the reference data distribution $R$, in the sense that 
$$\Delta(R|_{D(x)},S|_{D(x)})<\frac{1}{2}$$
See Figure \ref{fig:data_plag_index}(a) as an example.

\subsection{Distance to Closest Records (DCR)}

The Distance to Closest Records (DCR) technique was developed to identify identical matches between synthetic and training data \cite{park2018data, lu2019empirical}. However, it is essential to note that detecting identical matches and identifying data-copying are distinct tasks. While distance measures are appropriate for finding identical matches, they are not as effective for detecting data-copying, which is more appropriately viewed as a proportional measure. As discussed in Section 3.2 of \cite{platzer2021holdout}, a DCR of zero signifies an identical match, but a non-zero DCR does not capture the extent of personal information disclosure. This inherent flaw renders DCR an unsuitable measure of synthetic data privacy, as it does not accurately assess privacy risk. Our proposed Data Plagiarism Index (DPI) provides a more effective metric for synthetic data privacy by quantifying privacy risk. We utilize DPI to construct a Membership Attack, whose results reveal the privacy risks of synthetic data. See Section \ref{subsec:DPIMIA}.

\subsection{Membership Inference Attacks as Privacy Auditors}

Membership Inference Attacks (MIA) for synthetic data aim to determine if a given sample was a member of the original training set used to train a generative model. Consider a random variable $X$ defined on the domain $\mathcal{X}$, following a probability distribution $p_X(X)$. Let $D_{train}$ represent a training dataset consisting of independently sampled observations from $p_X(X)$. A generative model $G$ is then trained on $D_{train}$ and used to produce a synthetic dataset $D_{syn}$.
An attacker model, $\mathcal{A}: X \to {0, 1}$, has access to the synthetic dataset $D_{syn}$, a test data point $x^*$ drawn from the same distribution as $X$, and depending on the threat model, potentially other information. The attacker's objective is to determine whether the test point $x^*$ belongs to the original training set $D_{train}$. An ideal attacker would output $\mathcal{A}(x^*) = 1$ if $x^* \in D_{train}$, and $0$ otherwise. By considering privacy in this framework, MIAs allow for the practical evaluation of the risk of synthetic data release, relative to an adversarial scenario \cite{shroki}.

In this paper, we study a special case of Membership Inference Attacks in which the attacker has access to an additional reference dataset $D_{ref}$ sampled from same distribution as $D_{train}$ but not used in the training of $G$. There are a variety of both practical and theoretical reasons for the inclusion of $D_{ref}$ in the attack. First, $D_{ref}$ represents a worse-case scenario in which the attacker has access to the most information possible in which to build an attack which makes this situation important to study from a conservative, privacy conscious data publisher's perspective. Second, it is a real scenario that particularly tabular-orientated data publishers face in that $D_{ref}$ can be a leaked dataset, historic data, data from a competitor, or even a dataset built from publicly available data. Lastly, the inclusion of $D_{ref}$ is theoretically motivated from the growing body of literature involved with testing for data-copying in a generative model in that in order most frameworks contextualize model miss-specification between $D_{train}$ and $D_{syn}$ as being relative to a holdout dataset \cite{meehan2020three,bhattacharjee2023data, platzer2021holdout}.

\section{Measuring Data-Copying Misbehavior}
\label{sec:method}

\begin{figure}[t]
    \centering
    \includegraphics[width = 0.95\linewidth]{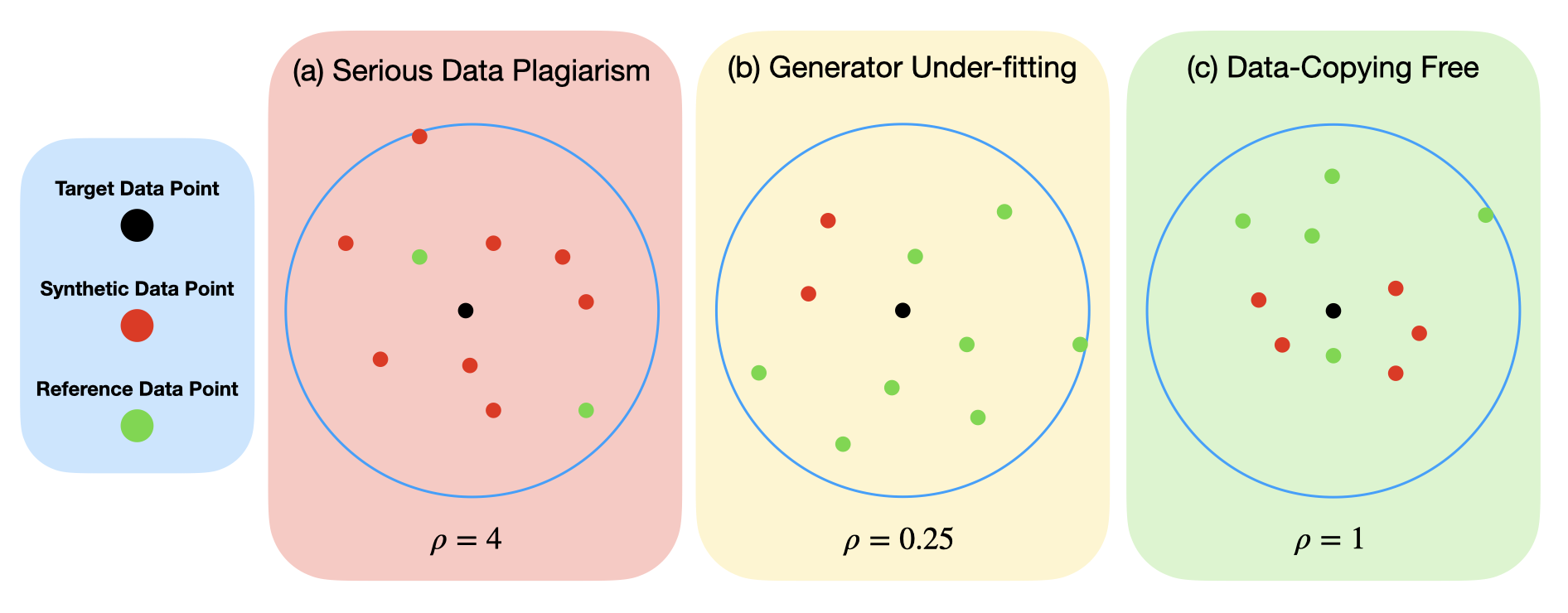}
    \caption{Data Plagiarism Index (DPI): We propose a novel privacy metric named Data Plagiarism Index (DPI). For each target data point (black point), we calculate the Data Plagiarism Index $\rho$ by first construct a k-nearest neighborhood (blue circle) around the target data point on the space with reference data points (green points) and synthetic data points (red points). The Data Plagiarism Index $\rho$ is defined simply as the ratio of number of  synthetic data points to the number of reference data points. See Sec. \ref{subsec:Data_Plag_Index} for whole details.}
\label{fig:data_plag_index}
\end{figure}

\subsection{Data Plagiarism Index (DPI)}
\label{subsec:Data_Plag_Index}

We propose a novel privacy metric called \textbf{Data Plagiarism Index (\texttt{DPI})}, as illustrated in Figure \ref{fig:data_plag_index}. For each training data point $x$, we calculate the Data Plagiarism Index $\rho$ by first constructing a K-Nearest Neighborhood $D(x)$ around $x$ on the space with reference data points and synthetic data points. The Data Plagiarism Index $\rho$ is defined simply as the ratio of number of synthetic data points to the number of reference data points; that is 
\begin{equation}\label{eq:data_plag_index}
\rho(x) \equiv \frac{\sum_{x_i \in D(x)}I(x_{i} \in S)}{\sum_{x_i \in D(x)}I(x_{i} \in R)}
\end{equation}

The DPI ranges from 0 to infinity, capturing a spectrum of data generation scenarios:
\begin{itemize}[leftmargin=*]
\item \textbf{DPI = 0}: Indicates no synthetic data points in the neighborhood, suggesting potential \textit{under-fitting} by the generative model.
\item \textbf{DPI = 1}: Represents an equal number of synthetic and reference data points, implying no data plagiarism or under-fitting, signifying a balanced data generation process.
\item \textbf{DPI $>$ 1}: Highlights a greater presence of synthetic data points compared to reference data points, with increasing values pointing towards significant \textit{data plagiarism}.
\end{itemize}
This comprehensive range allows the DPI to effectively capture various data generation scenarios, providing critical insights into the quality and privacy of synthetic datasets.

We provide 3 toy examples to help better understand how to interpret DPI values. In each example, consider the $K=10$ nearest neighborhood for the target data point $x$, denoted by $D(x) = \{x_1, \cdots, x_n\}$. Each element $x_i$ in the neighborhood may come from the synthetic dataset $S$ or reference dataset $R$:

\textbf{Toy Example 1 (Data Plagiarism, Figure \ref{fig:data_plag_index}.(a))}. Say 8 elements come from the synthetic dataset and 2 elements come from the reference dataset. Then the Data Plagiarism Index $\rho(x) = 8/2 = 4$, which means the number of synthetic data points is 4 times that of the reference data points! This marks serious data-copying behavior from that generative model. 

\textbf{Toy Example 2 (Generator Under-fitting, Figure \ref{fig:data_plag_index}.(b))}. In this situation, say 2 elements come from the synthetic dataset and 8 elements from the reference dataset. Then the Data Plagiarism Index $\rho(x) = 2/8 = 1/4$, which means the number of synthetic data points is 25\% of that of the reference data points! This marks little data-copying behavior, but also suggests a potential underfitting issue around the target data point $x$. 

\textbf{Toy Example 3 (Data-Copying Free, Figure \ref{fig:data_plag_index}.(c)).} Lastly, say there are 5 items from the synthetic dataset and 5 from the reference dataset. The Data Plagiarism Index is then $\rho(x) = 5/5 = 1$, meaining the number of synthetic data points is equal to the number of reference data points! This marks no data plagiarism and no underfitting behavior around the target data point $x$. 
    
In summary, the Data Plagiarism Index offers a robust and intuitive metric for evaluating the balance between synthetic and reference data points within a specified neighborhood around a target data point. By calculating the ratio of synthetic to reference data points, the DPI effectively highlights critical instances of data plagiarism, under-fitting, or a balanced data generation process. This metric is invaluable for rigorously assessing the performance of generative models, ensuring data integrity, and proactively identifying potential issues in synthetic data generation.

\begin{figure}[t]
   \centering
    \includegraphics[width = 1.0\linewidth]{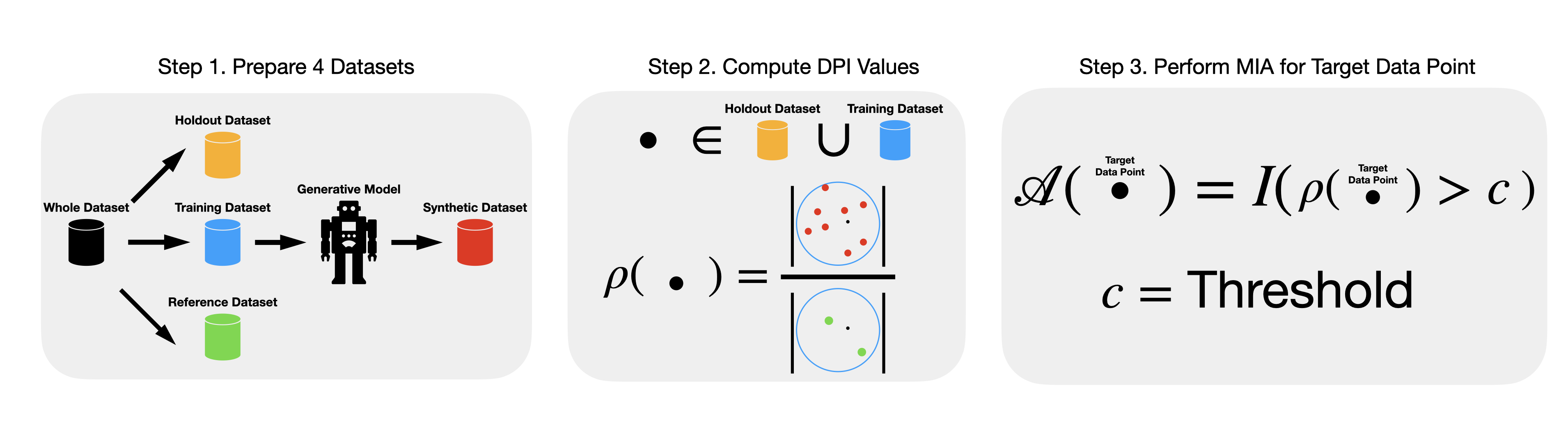}
    \caption{Data Plagiarism Index Membership Inference Attack (DPI MIA) See Sec. \ref{subsec:DPIMIA} for full details.}
\label{fig:data_plag_prober}
\end{figure}

\subsection{Data Plagiarism Membership Inference Attack}
\label{subsec:DPIMIA}

While the Data Plagiarism Index defined at Equation \eqref{eq:data_plag_index} provides insight into the local behavior of generative models, on its own it does not characterize the actual \textit{privacy risk} of data-copying in tabular generative models. We therefore show that a Membership Inference Attack can be derived from DPI which is designed to measure and then attack the local data-copying of generative models to evaluate the privacy risk. 

Note that, in the literature of similarity metrics (Sec. \ref{subsec:ref_similarity_metrics}), only the training dataset $D_{train}$ is compared to the synthetic $D_{syn}$ and reference $D_{ref}$ sets. In the MIA paradigm however (Sec. \ref{subsec:ref_MIA}), a fourth holdout dataset $D_{holdout}$ is taken and combined with the training dataset post generator training: $X = D_{train} \cup D_{holdout} $. The goal of an attacker $\mathcal{A}(x)$ is to effectively discriminate which set each record $x \in X$ originated from.

The Membership Inference Attack based on the Data Plagiarism Index can divided into three steps (See Figure \ref{fig:data_plag_prober}): 

\textbf{Step 1: Prepare 4 Datasets.} In the initial step, four distinct datasets are prepared to facilitate an attack construction. The foundation begins with the original dataset (denoted in black), which is divided into 3 equal sized sub-sets: the \textit{Training Dataset} (blue), the \textit{Holdout Dataset} (Yellow) and the \textit{Reference Dataset} (green). The Training Dataset is used to train the generative model, which subsequently generates the \textit{Synthetic Dataset} (red). In parallel, the Holdout Dataset serves as an independent test set, deliberately excluded from the training phase.

\textbf{Step 2: Compute DPI Values.} The second step involves the computation of the \textit{Data Plagiarism Index (DPI)} values defined at \eqref{eq:data_plag_index} for each data point within both the Holdout and Training datasets. This identifies data-copying in the local neighborhoods of test points from both datasets. The intuition here is that scores for test points from the Training set should theoretically be higher than the Holdout set if a model is vulnerable to copying data.

\textbf{Step 3: Perform MIA for Each Test Point} In Step 3, a \textit{Membership Inference Attack (MIA)} is executed to evaluate the privacy risk associated with a specific target data point. This uses the DPI score of the target data point, which is compared against a predefined threshold $c$. The attack can be written as: 
\begin{equation}\label{eq:DPMIA}
\mathcal{A}(\cdot) = I(\rho(\cdot) > c),    
\end{equation}
where $\mathcal{A}$ denotes the attack function, $\rho(\cdot)$ signifies the DPI value, and $I$ is an indicator function assessing if the DPI surpasses the threshold $c$. Should the DPI value of the target data point exceed $c$, it indicates that the data point is likely to have been part of the training data, thereby exposing a potential privacy breach.
In practice, MIAs are often benchmarked using AUCROC and so the choice of $c$ is largely irrelevant. Our implementation chooses $c$ as the median of $\rho(x)$ for $x$ for all test data.

The proposed DPI MIA defined at Equation \eqref{eq:DPMIA}, though straightforward, is a fast and easily interpretable attack, well-suited for auditing privacy in high-dimensional, mixed-type datasets. Indeed, as it only uses a K-Nearest Neighbor Search, it can be ran with linear or logarithmic time complexity (see \cite{osti_1443274}, \cite{international1989five}). Moreover, DPI MIA is compatible with various definitions of distance, making it versatile to many specific applications.

\section{Results}
\label{sec:results}

\subsection{Experiment Setup}

We are interested in studying how DPI compares with other Membership Inference Attacks, to what extent different tabular data generator architectures copy data, and if there are trends in the kind of data copied. To investigate these research questions, we benchmark a variety of model architectures and MIAs on the Adult Census dataset \cite{adult}. We provide descriptions of each model and MIA in appendix \ref{app:modeldescptions} and \ref{app:MIAdescriptions} respectively. In each experiment, Adult is randomly split into 3 equal sized training, holdout and reference sets where the generated synthetic size is also equal to these set. We replicate each experiment 5 times and visualize or report the means and standard deviations of the measures of interest where applicable. 

For DPI, $K$ has to be chosen as a hyperparameter and for data visualization purposes we plot the best performing DPI attack (Using L2 distance and $K$=20). We refer to appendix \ref{} for an ablation study of Section \ref{subsec:benchmarkDPI} containing plots with various distance metrics and $K$ levels but in practice, DPI is fairly stable in its results regardless of $K$ and we observed no extreme changes in behavior based on these hyperparameters.

\begin{figure*}
    \centering
    \includegraphics[width=\linewidth]{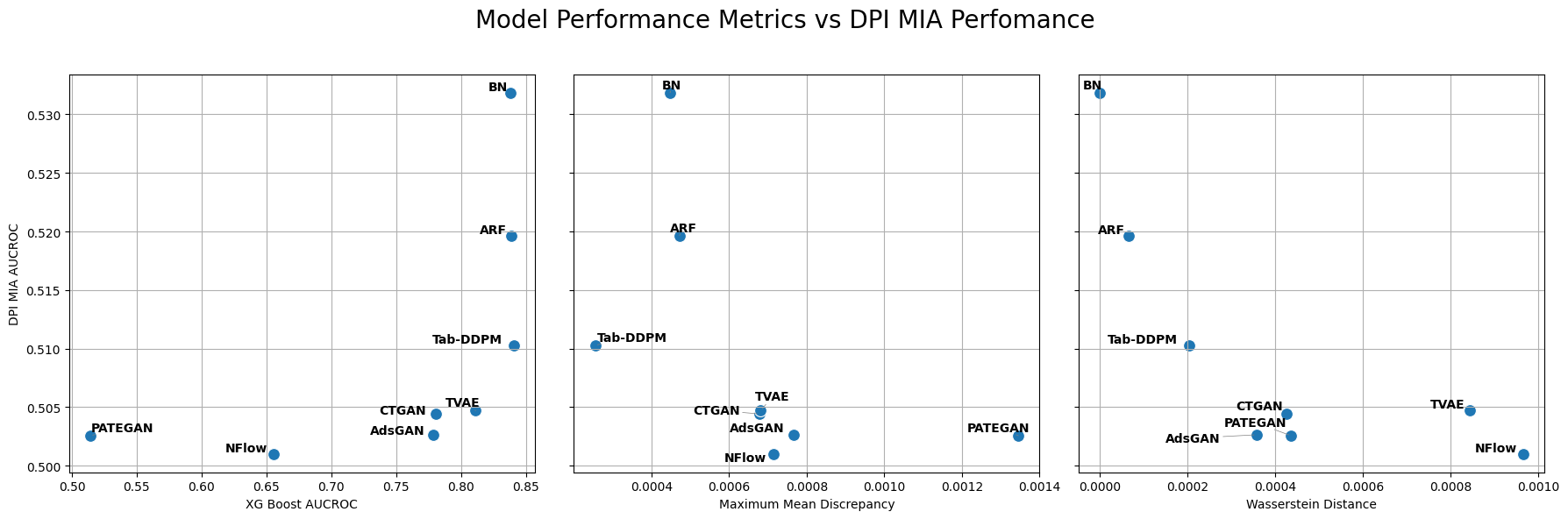}
    \caption{Classifier AUCROC, Maximum Mean Discrepancy, and Wasserstein Distance plotted with corresponding DPI MIA AUCROC for various common architectures. Interestingly, Bayesian Network, Adversarial Random Forest, and Tab-DDPM outperform other models in these performance metrics but have higher privacy risk. See Sec. \ref{subsec:data_copy_tabu_data_generator} for full details.}
    \label{fig:model_perf_v_mia}
\end{figure*}
\subsection{Data-Copying in Tabular Data Generators}
\label{subsec:data_copy_tabu_data_generator}

We first wish to confirm that there is a privacy risk to data copying in common tabular generators. Here, we benchmark a wide range of tabular generator strategies including: CTGAN and TVAE \cite{Xu2019ModelingTD}, Normalizing Flows (NFlow) \cite{durkan2019neural}, Bayesian Network (BN) \cite{Ankan2015}, Adversarial Random Forests (ARF) \cite{pmlr-v206-watson23a}, Tab-DDPM \cite{tabddpm}, PATEGAN \cite{yoon2018pategan}, and Ads-GAN \cite{PMID:32167919}. We evaluate these models on several metrics of synthetic data quality: the Maximum Mean Discrepancy between the training and generated data (lower is better), the Wasserstein distance between the marginals of the training and generated data (lower is better), and finally the AUCROC of an XGBoost \cite{xgboost} classifier trained on the synthetic set and tasked to predict on a holdout set (higher is better). We deploy the DPI MIA attack as described in Section \ref{subsec:DPIMIA} and report its AUCROC.

We visualize the mean values of utility metrics plotted against the AUCROC of DPI MIA in Figure \ref{fig:model_perf_v_mia}. Very interestingly, we observe that there is a clear trend between model performance and privacy risk. This suggests that data-copying could be a reason for why these models perform better, but additional research would need to be conducted to see if this is a causal or correlative relationship. 
\begin{figure}
        \includegraphics[width=\linewidth]{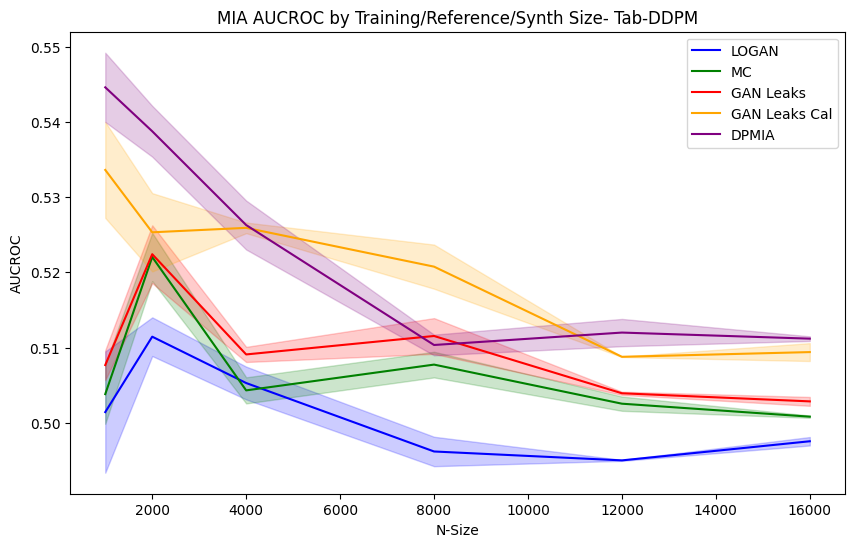}
        \caption{MIA AUCROC Benchmarks by Training Set Size on Tab-DDPM. This shows that DPI MIA is more effective than other existing MIAs described at Sec. \ref{subsec:ref_MIA}. 
        See Sec. \ref{subsec:benchmarkDPI} for full details.}
        \label{fig:DDPM_MIA}
    \end{figure}

\subsection{Properties of Training Data with High DPI Scores}
\label{subsec:Train_Data_High_DPI}

We also investigated the training data egregiously copied according to DPI. Here, we identified the top 1\% highest scored training data based on synthetic data generated by Tab-DDPM \cite{tabddpm} as it is widely used as a benchmark in the current tabular generation literature \cite{autodiff,stasy,tabsyn}. We visualize these data in Figure \ref{fig:tsne} as a t-SNE plot \cite{Maaten2008VisualizingDU} with the top 1\% highly scored indexes by DPI in red. Worryingly, DPI identifies an outlier region of the distribution as being subject to this extreme top percentile data-copying (the bottom right of the plot). When analyzing these observation themselves, we found that an extreme majority were all examples of married, white, middle aged, high capital gains, private industry, respondents making >50k in income. In the algorithmic fairness literature that often uses Adult in benchmarking, this is considered to be a minority but very privileged class \cite{mehrabi2022survey}. This suggests that not only is there a technical and privacy concern with data-copying, but it could also exacerbate unfairness.

\begin{figure*}
    \centering
    \includegraphics[width=\linewidth]{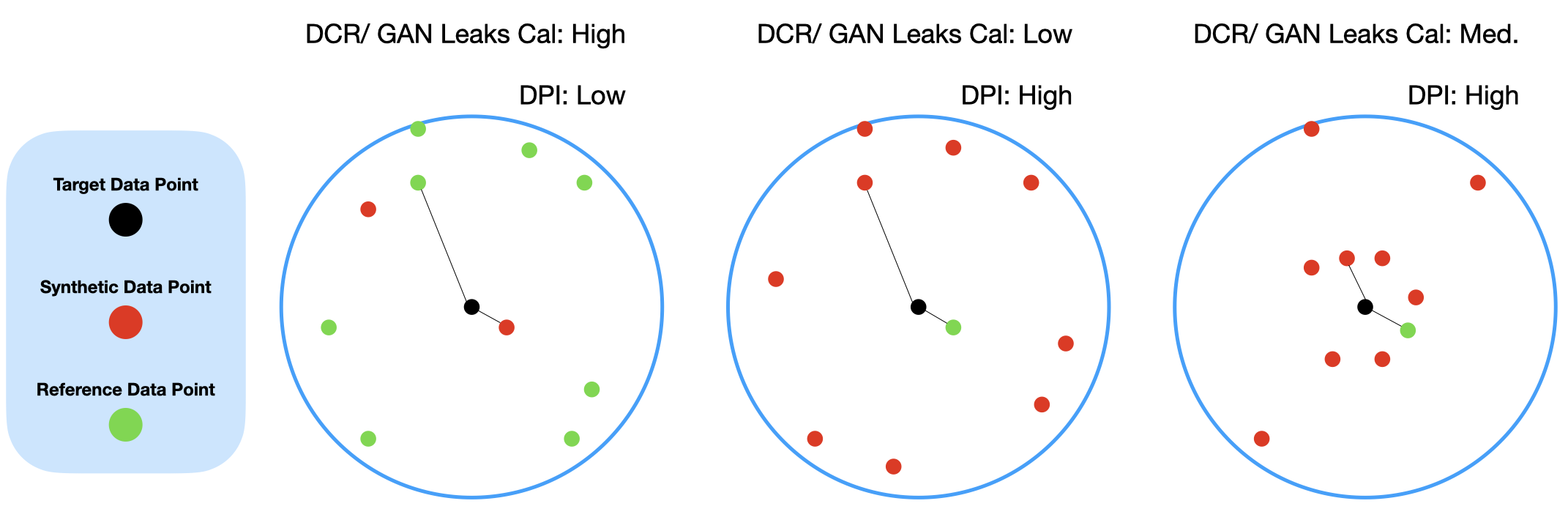}
    \caption{Distance Based Metrics/ MIAs of Overfitting vs DPI. DCR and GAN Leaks Calibrated evaluate overfitting based on a difference in relative distances whereas DPI evaluates data-copying based on the larger local neighborhood. This means that DPI evaluates data-copying differently than competing methods, explaining why predictions scores are not highly correlated (See Table \ref{tab:pcor}). See Sec. \ref{subsec:method_diff_DPI} for full details.}
    \label{fig:data_copying_differences}
\end{figure*}

\subsection{Benchmarking the Privacy Risk of DPI}
\label{subsec:benchmarkDPI}
Lastly, we evaluate DPI in relation to other MIAs in order to understand its efficacy as an MIA strategy. We evaluate the black box attacks of MC \cite{Hilprecht2019MonteCA}, and GAN Leaks \cite{ganleaks} as well as the calibrated attacks of LOGAN \cite{Hayes2017LOGANMI}, GAN Leaks Calibrated \cite{ganleaks} and DPI. We also benchmarked DOMIAS \cite{vanbreugel2023membership}, but we found that its density estimation failed to converge and its results were always the equivalent of random guess and therefore we do not formally report it. Again, we evaluate these methods on Tab-DDPM \cite{tabddpm}. In Figure \ref{fig:DDPM_MIA}, we display the results for each of these methods across various $D_{train}$/ $D_{ref}$/ $D_{syn}$ sizes. For simplicity, each dataset is equal in size to its other. 

\begin{table}[]
\caption{AUCROC for DPI and GAN Leaks Calibrated with Corresponding Pearson Correlations for the Predicted Scores. This tables shows that the proposed DPI identifies different types of data-copying than the one identified by the existing MIA attack. See Sec. \ref{subsec:benchmarkDPI} for whole details.}
\label{tab:pcor}
\begin{center}
\begin{tabular}{l|ccc}
\toprule
\multirow{2}{*}{\textbf{Model}} & \multicolumn{3}{c}{\textbf{Metric}}                                                     \\ \cmidrule(l){2-4}
                                & \textbf{DPI MIA} & \textbf{GAN Leaks CAL} & \textbf{$r$} \\ \midrule
Tab-DPDM                        & 0.510±0.004             & 0.507±0.003                   & 0.404±0.012                  \\ 
BN                              & 0.532±0.005             & 0.623±0.000                   & 0.268±0.020                  \\
ARF                             & 0.520±0.000             & 0.524±0.001                   & 0.339±0.006                  \\
AdsGAN                          & 0.503±0.003             & 0.501±0.003                   & 0.467±0.039                  \\
CTGAN                           & 0.504±0.003             & 0.501±0.005                   & 0.444±0.063                  \\
NFlow                           & 0.501±0.006             & 0.500±0.000                   & 0.482±0.023                  \\
PATEGAN                         & 0.503±0.003             & 0.502±0.003                   & 0.516±0.018                  \\
TVAE                            & 0.505±0.005             & 0.502±0.003                   & 0.466±0.023                  \\
\bottomrule
\end{tabular}
\end{center}
\end{table}

Overall, DPI is competitive with other Membership Inference Attacks compatible with these data, dominating at most testing sizes. Interestingly, GAN Leaks Calibrated performs similarly or better than DPI. However, GAN Leaks Calibrated scores test data very differently than DPI, being based on a difference in the relative distances of the closest synthetic and reference points. In Table \ref{tab:pcor} we show that for various models, while the performance of GAN Leaks Calibrated and DPI can differ, the Pearson Correlation between their scores is relatively low. This implies that while these measures are correlated, DPI is picking up on its specific data-copying definition. Indeed, the motivation of this work is not to create a State of the Art Membership Inference Attack for all architectures, but rather to characterize the unique risk DPI implies.

\section{Discussions}
\subsection{Implications for Privacy and Fairness in Synthetic Data}

Data Plagiarism Index provides evidence that popular tabular generative models can exhibit risky data-copying behavior. This includes leaking information and favoring specific sub-class outliers in the distribution of the training data. Thus, DPI can be used as a tool to audit and study model behavior. It should be noted however, that while DPI can show synthetic data are not private, it cannot prove the opposite: that particular synthetic data are private. For example, different MIAs may have various levels of success if their method targets different attributes of the synthetic data. This is proved by Table \ref{tab:pcor} where DPI and GAN Leaks Calibrated predicted scores, while correlated, were not perfectly aligned. 

On top of model auditing, DPI can be connected to Differential Privacy applications \cite{dwork2006calibrating}. Here, DPI can be used to 
evaluate the practical lower bound of the privacy parameter $\epsilon$. For example, one approach is to select a pair of neighboring training datasets $D_1$ and $D_2$ and produce corresponding synthetic datasets $\tilde{D}_1$ and $\tilde{D}_2$ with a generative model. With DPI, we can then create corresponding data copying score distributions to find a decision rule that, given an unknown synthetic dataset $\tilde{D}$, identifies whether its source was $D_1$ or $D_2$. If the decision rule's true positive rate is $\alpha$ and the false negative rate is $\beta$, based on the remark after Theorem 1 in \cite{houssiau2022tapas}, the privacy budget's lower bound can be expressed as $\epsilon \geq \log \max \left\{\frac{\alpha}{1-\beta}, \frac{\beta}{1-\alpha}\right\}$.
This approach quantifies the minimum differential privacy level that the generative model upholds.

\subsection{Methodological Differences in DPI}
\label{subsec:method_diff_DPI}

DPI provides a new geometric definition for data-copying in the context of an available reference set and uniquely attacks this attribute relative to other MIAs. In Figure \ref{fig:data_copying_differences}, we show a variety of scenarios in which hypothetical test data points are plotted with their closest synthetic and reference set neighbors. We show that under the Distance to Closest Record (DCR) metric and GAN Leaks Calibrated MIA \cite{ganleaks} (which in many respects is the MIA version of DCR) understanding of data-copying, certain scenarios would be classified differently where they would label a positive instance of overfitting based on extreme differences in the distances of the nearest synthetic and reference points whereas DPI labels it based off of extreme differences in proportions. Thus, DPI evaluates data-copying in a fundamentally different way than DCR/ GAN Leaks Calibrated and provides additional insight into how the training and synthetic data are distributed.

\subsection{Challenges in Tabular Data Generation}
A key challenge of generative modelling with tabular data is the unstructured, high dimensional, mixed type nature of most datasets \cite{Xu2019ModelingTD}. This poses a challenge for newer results on model data-copying and Membership Inference attacks that focus on density estimation. \cite{bhattacharjee2023data} for example proposes a scheme of comparing local densities of training data and synthetic data but does not frame their work from an MIA perspective. Similarly, they provide a proof that their method is not effective on non-smooth distributions that are characteristic in the tabular domain. DOMIAS \cite{vanbreugel2023membership} evaluates overfitting as an MIA by comparing a test point to the probability densities of the synthetic and reference distributions. They propose two options for estimating these densities in using a Gaussian Kernel Density Estimator and a deep learning method called Block Neural Autoregressive Flow (BNAF) \cite{bnaf19}. We found however that these methods have difficulty converging with high dimensional, mixed type datasets. Indeed, the highest dimensional tabular dataset that was benchmarked in DOMIAS was a private healthcare dataset that when one-hot-encoded was 35 columns. Adult when similarly pre-processed is 109 columns. All of our experiments with DOMIAS failed to converge, leaving its results as being the equivalent of random guess. Indeed the authors note a limitation of the work is its reliance on BNAF in that it can take several hours to train. This motivates this paper to consider a computationally easier paradigm of analyzing local neighborhoods around test points.

\section{Conclusion and Future Work}
In this paper, we propose a novel measure of data-copying and connect it to the Membership Inference Attack literature for tabular generative models. This allows the unique study of how local data-copying contributes to risks in the trustworthiness of these generators. Models that perform well in generating data with high measures of utility tend to copy training data more than models of a lower quality and thus have higher privacy risk profiles. Similarly, we have shown that Tab-DDPM, a highly cited and studied architecture egregiously copies outlier training data of privileged sub-classes in a widely used fairness benchmarking dataset. This indicates that data-copying can effect a variety of concepts, such as fairness, used to evaluate the trustworthiness of generative models.

Data Plagiarism Index motivates a variety of directions for future work. The disparate nature of the data-copying literature necessitates a broader theoretical framework in which to connect data-copying to privacy. Similarly, this work has shown that data-copying can affect different axis' in which to evaluate the trustworthiness of models. It would be interesting to further explore if it also effects other aspects of Trustworthy AI such as robustness, interpretability, and reliability. Lastly, DPI can be applied to Differential Privacy Auditing where it can be used to evaluate sharper privacy lower bounds.

\clearpage

%\clearpage
\baselineskip=13pt
\bibliographystyle{plain}
\nocite{*}
\bibliography{ref}

\begin{thebibliography}{10}

\bibitem{ims1}
Mostly AI.
\newblock Truly anonymous synthetic data -- evolving legal definitions and technologies (part ii).
\newblock 2020.

\bibitem{ims2}
Mostly AI.
\newblock How to implement data privacy? a conversation with klaudius kalcher.
\newblock 2021.

\bibitem{alaa2022faithful}
Ahmed Alaa, Boris Van~Breugel, Evgeny~S Saveliev, and Mihaela van~der Schaar.
\newblock How faithful is your synthetic data? sample-level metrics for evaluating and auditing generative models.
\newblock In {\em International Conference on Machine Learning}, pages 290--306. PMLR, 2022.

\bibitem{Ankan2015}
Ankur Ankan and Abinash Panda.
\newblock pgmpy: Probabilistic graphical models using python.
\newblock In {\em Proceedings of the Python in Science Conference}, SciPy. SciPy, 2015.

\bibitem{adult}
Barry Becker and Ronny Kohavi.
\newblock {Adult}.
\newblock UCI Machine Learning Repository, 1996.
\newblock {DOI}: https://doi.org/10.24432/C5XW20.

\bibitem{bhattacharjee2023data}
Robi Bhattacharjee, Sanjoy Dasgupta, and Kamalika Chaudhuri.
\newblock Data-copying in generative models: a formal framework.
\newblock In {\em International Conference on Machine Learning}, pages 2364--2396. PMLR, 2023.

\bibitem{CaseBerg:01}
George Casella and Roger Berger.
\newblock {\em Statistical Inference}.
\newblock {Duxbury Resource Center}, June 2001.

\bibitem{ganleaks}
Dingfan Chen, Ning Yu, Yang Zhang, and Mario Fritz.
\newblock Gan-leaks: A taxonomy of membership inference attacks against generative models.
\newblock In {\em Proceedings of the 2020 ACM SIGSAC Conference on Computer and Communications Security}, CCS ’20. ACM, October 2020.

\bibitem{xgboost}
Tianqi Chen and Carlos Guestrin.
\newblock Xgboost: A scalable tree boosting system.
\newblock In {\em Proceedings of the 22nd ACM SIGKDD International Conference on Knowledge Discovery and Data Mining}, KDD '16, page 785–794, New York, NY, USA, 2016. Association for Computing Machinery.

\bibitem{cheng2024downstream}
Yinan Cheng, Chi-Hua Wang, Vamsi~K Potluru, Tucker Balch, and Guang Cheng.
\newblock Downstream task-oriented generative model selections on synthetic data training for fraud detection models.
\newblock {\em arXiv preprint arXiv:2401.00974}, 2024.

\bibitem{bnaf19}
Nicola De~Cao, Ivan Titov, and Wilker Aziz.
\newblock Block neural autoregressive flow.
\newblock {\em 35th Conference on Uncertainty in Artificial Intelligence (UAI19)}, 2019.

\bibitem{durkan2019neural}
Conor Durkan, Artur Bekasov, Iain Murray, and George Papamakarios.
\newblock Neural spline flows, 2019.

\bibitem{dwork2006calibrating}
Cynthia Dwork, Frank McSherry, Kobbi Nissim, and Adam Smith.
\newblock Calibrating noise to sensitivity in private data analysis.
\newblock In {\em Theory of Cryptography: Third Theory of Cryptography Conference, TCC 2006, New York, NY, USA, March 4-7, 2006. Proceedings 3}, pages 265--284. Springer, 2006.

\bibitem{osti_1443274}
Jerome~H. Friedman, Jon~Louis Bentley, and Raphael~Ari Finkel.
\newblock An algorithm for finding best matches in logarithmic expected time.
\newblock {\em ACM Transactions on Mathematical Software}, 3(3), 7 1976.

\bibitem{ganev2023inadequacy}
Georgi Ganev and Emiliano~De Cristofaro.
\newblock On the inadequacy of similarity-based privacy metrics: Reconstruction attacks against "truly anonymous synthetic data'', 2023.

\bibitem{guillaudeux2023patient}
Morgan Guillaudeux, Olivia Rousseau, Julien Petot, Zineb Bennis, Charles-Axel Dein, Thomas Goronflot, Matilde Karakachoff, Sophie Limou, Nicolas Vince, Matthieu Wargny, and Pierre-Antoine Gourraud.
\newblock Patient-centric synthetic data generation, no reason to risk re-identification in the analysis of biomedical pseudonymised data.
\newblock 05 2022.

\bibitem{Hayes2017LOGANMI}
Jamie Hayes, Luca Melis, George Danezis, and Emiliano~De Cristofaro.
\newblock Logan: Membership inference attacks against generative models.
\newblock {\em Proceedings on Privacy Enhancing Technologies}, 2019:133 -- 152, 2017.

\bibitem{Hilprecht2019MonteCA}
Benjamin Hilprecht, Martin H{\"a}rterich, and Daniel Bernau.
\newblock Monte carlo and reconstruction membership inference attacks against generative models.
\newblock {\em Proceedings on Privacy Enhancing Technologies}, 2019:232 -- 249, 2019.

\bibitem{houssiau2022tapas}
Florimond Houssiau, James Jordon, Samuel~N Cohen, Owen Daniel, Andrew Elliott, James Geddes, Callum Mole, Camila Rangel-Smith, and Lukasz Szpruch.
\newblock Tapas: a toolbox for adversarial privacy auditing of synthetic data.
\newblock {\em arXiv preprint arXiv:2211.06550}, 2022.

\bibitem{hsieh2024improve}
Din-Yin Hsieh, Chi-Hua Wang, and Guang Cheng.
\newblock Improve fidelity and utility of synthetic credit card transaction time series from data-centric perspective.
\newblock {\em arXiv preprint arXiv:2401.00965}, 2024.

\bibitem{international1989five}
International Computer~Science Institute and S.M. Omohundro.
\newblock {\em Five Balltree Construction Algorithms}.
\newblock Technical report (International Computer Science Institute). International Computer Science Institute, 1989.

\bibitem{stasy}
Jayoung Kim, Chaejeong Lee, and Noseong Park.
\newblock {ST}asy: Score-based tabular data synthesis.
\newblock In {\em The Eleventh International Conference on Learning Representations}, 2023.

\bibitem{tabddpm}
Akim Kotelnikov, Dmitry Baranchuk, Ivan Rubachev, and Artem Babenko.
\newblock Tabddpm: Modelling tabular data with diffusion models, 2022.

\bibitem{li2021online}
Yuantong Li, Chi-Hua Wang, and Guang Cheng.
\newblock Online forgetting process for linear regression models.
\newblock In {\em International Conference on Artificial Intelligence and Statistics}, pages 217--225. PMLR, 2021.

\bibitem{liu2023tabular}
Tongyu Liu, Ju~Fan, Guoliang Li, Nan Tang, and Xiaoyong Du.
\newblock Tabular data synthesis with generative adversarial networks: design space and optimizations.
\newblock {\em The VLDB Journal}, 33(2):255–280, aug 2023.

\bibitem{liu2022utility}
Yucong Liu, Chi-Hua Wang, and Guang Cheng.
\newblock On the utility recovery incapability of neural net-based differential private tabular training data synthesizer under privacy deregulation.
\newblock {\em arXiv preprint arXiv:2211.15809}, 2022.

\bibitem{lu2019empirical}
Pei-Hsuan Lu, Pang-Chieh Wang, and Chia-Mu Yu.
\newblock Empirical evaluation on synthetic data generation with generative adversarial network.
\newblock In {\em Proceedings of the 9th International Conference on Web Intelligence, Mining and Semantics}, WIMS2019, New York, NY, USA, 2019. Association for Computing Machinery.

\bibitem{meehan2020three}
Casey Meehan, Kamalika Chaudhuri, and Sanjoy Dasgupta.
\newblock A three sample hypothesis test for evaluating generative models.
\newblock In {\em International Conference on Artificial Intelligence and Statistics}, pages 3546--3556. PMLR, 2020.

\bibitem{Meeus_2024}
Matthieu Meeus, Florent Guepin, Ana-Maria Creţu, and Yves-Alexandre de~Montjoye.
\newblock {\em Achilles’ Heels: Vulnerable Record Identification in Synthetic Data Publishing}, page 380–399.
\newblock Springer Nature Switzerland, 2024.

\bibitem{mehrabi2022survey}
Ninareh Mehrabi, Fred Morstatter, Nripsuta Saxena, Kristina Lerman, and Aram Galstyan.
\newblock A survey on bias and fairness in machine learning, 2022.

\bibitem{park2018data}
Noseong Park, Mahmoud Mohammadi, Kshitij Gorde, Sushil Jajodia, Hongkyu Park, and Youngmin Kim.
\newblock Data synthesis based on generative adversarial networks.
\newblock {\em arXiv preprint arXiv:1806.03384}, 2018.

\bibitem{Park_2018}
Noseong Park, Mahmoud Mohammadi, Kshitij Gorde, Sushil Jajodia, Hongkyu Park, and Youngmin Kim.
\newblock Data synthesis based on generative adversarial networks.
\newblock {\em Proceedings of the VLDB Endowment}, 11(10):1071–1083, June 2018.

\bibitem{parker2006case}
Elizabeth~S Parker, Larry Cahill, and James~L McGaugh.
\newblock A case of unusual autobiographical remembering.
\newblock {\em Neurocase}, 12(1):35--49, 2006.

\bibitem{platzer2021holdout}
Michael Platzer and Thomas Reutterer.
\newblock Holdout-based empirical assessment of mixed-type synthetic data.
\newblock {\em Frontiers in big Data}, 4:679939, 2021.

\bibitem{synthcity}
Zhaozhi Qian, Bogdan-Constantin Cebere, and Mihaela van~der Schaar.
\newblock Synthcity: facilitating innovative use cases of synthetic data in different data modalities, 2023.

\bibitem{sablayrolles2019white}
Alexandre Sablayrolles, Matthijs Douze, Cordelia Schmid, Yann Ollivier, and Herv{\'e} J{\'e}gou.
\newblock White-box vs black-box: Bayes optimal strategies for membership inference.
\newblock In {\em International Conference on Machine Learning}, pages 5558--5567. PMLR, 2019.

\bibitem{shroki}
R.~Shokri, M.~Stronati, C.~Song, and V.~Shmatikov.
\newblock Membership inference attacks against machine learning models.
\newblock In {\em 2017 IEEE Symposium on Security and Privacy (SP)}, pages 3--18, Los Alamitos, CA, USA, may 2017. IEEE Computer Society.

\bibitem{solatorio2023realtabformer}
Aivin~V Solatorio and Olivier Dupriez.
\newblock Realtabformer: Generating realistic relational and tabular data using transformers.
\newblock {\em arXiv preprint arXiv:2302.02041}, 2023.

\bibitem{autodiff}
Namjoon Suh, Xiaofeng Lin, Din-Yin Hsieh, Mehrdad Honarkhah, and Guang Cheng.
\newblock Autodiff: combining auto-encoder and diffusion model for tabular data synthesizing.
\newblock In {\em NeurIPS 2023 Workshop on Synthetic Data Generation with Generative AI}, 2023.

\bibitem{tao2024discriminative}
Lan Tao, Shirong Xu, Chi-Hua Wang, Namjoon Suh, and Guang Cheng.
\newblock Discriminative estimation of total variation distance: A fidelity auditor for generative data.
\newblock {\em arXiv preprint arXiv:2405.15337}, 2024.

\bibitem{vanbreugel2023membership}
Boris van Breugel, Hao Sun, Zhaozhi Qian, and Mihaela van~der Schaar.
\newblock Membership inference attacks against synthetic data through overfitting detection, 2023.

\bibitem{Maaten2008VisualizingDU}
Laurens van~der Maaten and Geoffrey~E. Hinton.
\newblock Visualizing data using t-sne.
\newblock {\em Journal of Machine Learning Research}, 9:2579--2605, 2008.

\bibitem{wang2024badgd}
Chi-Hua Wang and Guang Cheng.
\newblock Badgd: A unified data-centric framework to identify gradient descent vulnerabilities.
\newblock {\em arXiv preprint arXiv:2405.15979}, 2024.

\bibitem{pmlr-v206-watson23a}
David~S. Watson, Kristin Blesch, Jan Kapar, and Marvin~N. Wright.
\newblock Adversarial random forests for density estimation and generative modeling.
\newblock In Francisco Ruiz, Jennifer Dy, and Jan-Willem van~de Meent, editors, {\em Proceedings of The 26th International Conference on Artificial Intelligence and Statistics}, volume 206 of {\em Proceedings of Machine Learning Research}, pages 5357--5375. PMLR, 25--27 Apr 2023.

\bibitem{Xu2019ModelingTD}
Lei Xu, Maria Skoularidou, Alfredo Cuesta-Infante, and Kalyan Veeramachaneni.
\newblock Modeling tabular data using conditional gan.
\newblock In {\em Neural Information Processing Systems}, 2019.

\bibitem{yale2019assessing}
Andrew Yale, Saloni Dash, Ritik Dutta, Isabelle Guyon, Adrien Pavao, and Kristin~P. Bennett.
\newblock Assessing privacy and quality of synthetic health data.
\newblock In {\em Proceedings of the Conference on Artificial Intelligence for Data Discovery and Reuse}, AIDR '19, New York, NY, USA, 2019. Association for Computing Machinery.

\bibitem{yoon2020anonymization}
Jinsung Yoon, Lydia~N Drumright, and Mihaela Van Der~Schaar.
\newblock Anonymization through data synthesis using generative adversarial networks (ads-gan).
\newblock {\em IEEE journal of biomedical and health informatics}, 24(8):2378--2388, 2020.

\bibitem{PMID:32167919}
Jinsung Yoon, Lydia~N Drumright, and Mihaela van~der Schaar.
\newblock Anonymization through data synthesis using generative adversarial networks (ads-gan).
\newblock {\em IEEE journal of biomedical and health informatics}, 24(8):2378—2388, August 2020.

\bibitem{yoon2018pategan}
Jinsung Yoon, James Jordon, and Mihaela van~der Schaar.
\newblock {PATE}-{GAN}: Generating synthetic data with differential privacy guarantees.
\newblock In {\em International Conference on Learning Representations}, 2019.

\bibitem{tabsyn}
Hengrui Zhang, Jiani Zhang, Zhengyuan Shen, Balasubramaniam Srinivasan, Xiao Qin, Christos Faloutsos, Huzefa Rangwala, and George Karypis.
\newblock Mixed-type tabular data synthesis with score-based diffusion in latent space.
\newblock In {\em The Twelfth International Conference on Learning Representations}, 2024.

\bibitem{zhao2021ctab}
Zilong Zhao, Aditya Kunar, Robert Birke, and Lydia~Y. Chen.
\newblock Ctab-gan: Effective table data synthesizing.
\newblock In Vineeth~N. Balasubramanian and Ivor Tsang, editors, {\em Proceedings of The 13th Asian Conference on Machine Learning}, volume 157 of {\em Proceedings of Machine Learning Research}, pages 97--112. PMLR, 17--19 Nov 2021.

\end{thebibliography}

\clearpage
\appendix

\section{Ablation study}
\label{app:Abalation Study}
\begin{figure}[H]
    \centering
    \includegraphics[width=\linewidth]{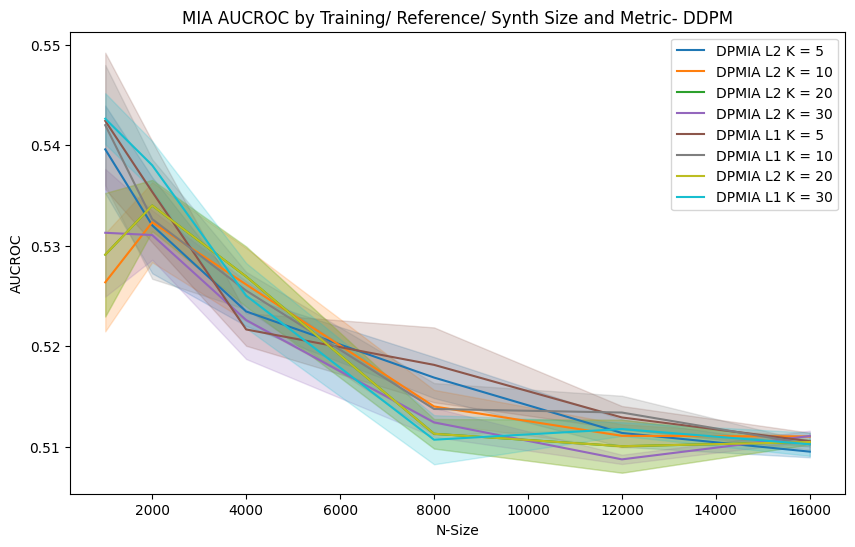}
    \caption{Ablation Results for Various Distance Measures and $K$ Size Choices.}
    \label{fig:Ablation}
\end{figure}

DPI requires a practitioner to specify a distance metric and $K$ number of nearest neighbors in order to be deployed. This presents a hyperparameter tuning problem as to what measure of distance should be used and how large the neighbors should be. We replicate the experiment from Section \ref{subsec:benchmarkDPI} but this time with common distance metrics (L1 and L2) as well as a variety of $K$ sizes (5, 10, 20, 30). We plot the means and standard deviations of the corresponding DPI MIA attacks in Figure \ref{fig:Ablation}. While the success of the attacks vary with lower sample sizes, each attack follows a clear trend with each, eventually seeing smaller deviations at the maximum sample sizes.

\section{Model Descriptions}
\label{app:modeldescptions}
In all experiments, we use the implementations of these models from the Python package Synthcity \cite{synthcity}. For benchmarking purposes we use the default hyperparameters for each model. A brief description for each model is as follows:

\textbf{CTGAN} \cite{Xu2019ModelingTD}: (Conditional Tabular Generative Adversarial Network) uses a GAN framework with conditional generator and discriminator to capture multi-modal distributions. It uses mode normalization to better learn mixed-type distributions.

\textbf{TVAE} \cite{Xu2019ModelingTD}: (Tabular Variational Auto-Encoder) is very similiar to CTGAN in its use of mode normalizing techniques, but rather than using a GAN architecture, instead employees A VAE.

\textbf{Normalizing Flows (NFlow)} \cite{durkan2019neural}: Normalizing flows transform a simple base distribution (e.g. Gaussian) into a more complex one matching the data by applying a sequence of invertible, differentiable mappings. 

\textbf{Bayesian Network (BN)} \cite{Ankan2015},: Bayesian Networks use a Directed Acyclic Graph to represent the joint probability distribution over variables as a product of marginal and conditional distributions. It then samples the empiric distributions estimated from the training dataset.

\textbf{Adversarial Random Forests (ARF)} \cite{pmlr-v206-watson23a}: ARFs extend the random forest model by adding an adversarial stage. Random forests generate synthetic samples which are scored against the real data by a discriminator network. This score is used to re-train the forests iteratively.

\textbf{Tab-DDPM }\cite{tabddpm}: Tabular Denoising Diffusion Probabilistic Model adapts the DDPM framework from image synthesis. It iteratively refines random noise into synthetic data by learning the data distribution through gradients of a classifier on partially corrupted samples with gaussian noise.

\textbf{PATEGAN} \cite{yoon2018pategan}: The PATEGAN model uses a neural encoder to map discrete tabular data into a continuous latent representation which is sampled from during generation by the GAN discriminator and generator pair.

\textbf{Ads-GAN} \cite{yoon2020anonymization}: Ads-GAN uses a GAN architecture for tabular synthesis but also adds an identifiability metric to increase its ability to not mimic training data.

\section{Membership Inference Attack Descriptions}
\label{app:MIAdescriptions}

A description of each of the Membership Inference Attacks referenced in the paper are as follows:

\textbf{LOGAN} \cite{Hayes2017LOGANMI}: LOGAN proposes a variety of MIA strategies. A black box version of their attack involves training a Generative Adversarial Network (GAN) on the synthetic dataset and using the discriminator to score test data. A calibrated version improves upon this by training a binary classifier to distinguish between the synthetic and reference dataset. In this paper we only benchmark the calibrated version.

\textbf{GAN Leaks/ GAN Leaks Calibrated} \cite{ganleaks}: GAN Leaks is a black box attack that scores test data based on a sigmoid score of the distance to the nearest neighbor in the synthetic dataset. GAN Leaks Calibrated improves on this with the inclusion of a reference set in which this distance is subtracted from the distance to the closest record in the reference set.

\textbf{MC} \cite{Hilprecht2019MonteCA}: MC is based on counting the amount of observations in the synthetic dataset that fall into the neighborhood of a test point (Monte Carlo Integration). However, they do not consider a reference dataset and the choice of distance for what to consider a neighborhood is a non-trivial hyperparameter to tune. 

\textbf{DOMIAS} \cite{vanbreugel2023membership}: DOMIAS is a calibrated attack which scores test data by performing density estimation on the synthetic and reference datasets to then calculate the probability ratio of the test data being from the synthetic vs reference distributions.

\end{document}